\documentclass[conference]{IEEEtran}
\IEEEoverridecommandlockouts
\usepackage{cite}
\usepackage{amsmath,amssymb,amsfonts}
\usepackage{algorithmic}
\usepackage{graphicx}
\usepackage{textcomp}
\usepackage{xcolor}
\usepackage{hyperref} 
\usepackage{balance} 

\usepackage{algorithm}
\usepackage{soul}
\usepackage{multirow}
\usepackage{array}

\usepackage{caption}
\usepackage{subcaption}

\def\BibTeX{{\rm B\kern-.05em{\sc i\kern-.025em b}\kern-.08em
    T\kern-.1667em\lower.7ex\hbox{E}\kern-.125emX}}
\begin{document}

\title{Spatial-Temporal Graph Representation Learning for Tactical Networks Future State Prediction}

\author{
    \IEEEauthorblockN{
        Junhua Liu\IEEEauthorrefmark{1}\IEEEauthorrefmark{2}, 
        Justin Albrethsen\IEEEauthorrefmark{1}, 
        Lincoln Goh\IEEEauthorrefmark{2}, 
        David Yau\IEEEauthorrefmark{1},
        Kwan Hui Lim\IEEEauthorrefmark{1}}
    \IEEEauthorblockA{\IEEEauthorrefmark{1}Singapore University of Technology and Design
    \\\{junhua\_liu, justin\_albrethsen, david\_yau, kwanhui\_lim\}@sutd.edu.sg}
    \IEEEauthorblockA{\IEEEauthorrefmark{2}Forth AI, \\\{j, lincoln\}@forth.ai}
}

\maketitle

\begin{abstract}
Resource allocation in tactical ad-hoc networks presents unique challenges due to their dynamic and multi-hop nature. Accurate prediction of future network connectivity is essential for effective resource allocation in such environments. In this paper, we introduce the Spatial-Temporal Graph Encoder-Decoder (STGED) framework for Tactical Communication Networks that leverages both spatial and temporal features of network states to learn latent tactical behaviors effectively. STGED hierarchically utilizes graph-based attention mechanism to spatially encode a series of communication network states, leverages a recurrent neural network to temporally encode the evolution of states, and a fully-connected feed-forward network to decode the connectivity in the future state. Through extensive experiments, we demonstrate that STGED consistently outperforms baseline models by large margins across different time-steps input, achieving an accuracy of up to 99.2\% for the future state prediction task of tactical communication networks. 


\end{abstract}



\section{Introduction}

Modern militaries require reliable communications on the tactical edge to support dynamic operations. Tactical communication networks (TCNs) extend 4G, 5G, and SATCOM communication capabilities to units without access to traditional network infrastructure. To this end, TCNs are typically formed as self-organizing mobile ad-hoc networks (MANETs) using high frequency (HF), very high frequency (VHF), and ultra high frequency (UHF) radio technologies \cite{7496568}. In TCNs it is critical to distribute information to support accurate and updated situational awareness. In these circumstances poor Quality of Service (QoS) such as dropped packets or high delays can lead to catastrophic military outcomes.

QoS is critical to TCNs and is also challenging to provide. One example of a QoS technique is RSVP which allows users to reserve networking resources for all nodes in an application stream's path \cite{4086811}. This relies on established routes and ample networking resources to guarantee QoS. However, TCNs often have low data-rates and routes consisting of multiple wireless hops with varying signal strengths \cite{8398729}. Furthermore, nodes are mobile across diverse terrain and may frequently break routes without warning, which makes the topology highly dynamic.  Even standard protocols like TCP can become ineffective when used in this kind of dynamic networks \cite{srinivasan_tcp-rta_2023}. TCNs are highly dynamic and routes often involve multiple hops, as such it is difficult to guarantee resources to every node involved in sending or routing critical applications. Forecasting future network connectivity in TCNs is a necessary step to enable resource reservation and QoS guarantees for priority network traffic \cite{8344739}. 




TCNs introduce unique challenges to traditional networking QoS solutions. To study these problems, the North Atlantic Treaty Organization (NATO) created the Anglova scenario \cite{peuhkuri_start_2022}. The Anglova scenario is designed to represent typical communication challenges in real TCNs and provide a standard experimentation environment for research  \cite{7496568}\cite{8842710}\cite{8398729}. The Anglova scenario includes sensor networks, naval task force, tank battalions, mechanized infantry, aerostat, UAVs, and medivac nodes. The scenario is organized into three vignettes to showcase different operational requirements. 

Within this scenario we look to predict future network connectivity as a precursor for future QoS work. Predicting future network connectivity can be formulated as a kind of link prediction problem, which has many successful solutions among a diverse group of domains including airline routes \cite{liu_link_2011}, social media \cite{adamic_friends_2003}, criminal networks \cite{berlusconi_link_2016}, and e-commerce \cite{huang_link_2005}. Link prediction has also been successful with certain MANETs using graph networks which can capture both spatial and temporal relationships that are present in dynamic networks \cite{shao_link_2022}. Existing work does attempt to predict future link quality in tactical networks, but their model assumes a simplified free space radio propagation model and does not account for any terrain effects \cite{10253464}. As part of the Anglova scenario, they provide radio path loss data based on a propagation model that incorporates complex terrain, making link prediction more complicated and realistic. 

This work aims to predict future connectivity in the realistic Anglova scenario. In summary, the main contributions include:

\begin{itemize}
\item \textbf{Datasets for TCNs:} Introduction of two benchmark datasets, Communication Network with Tactical Movement (CNTM) and Communication Network with Casual Movement (CNCM), based on the Anglova scenario to stimulate future TCN research~\footnote{available at: \url{https://github.com/junhua/STGED}}.
\item \textbf{STGED Framework:} Introduction of STGED framework that integrates Graph Neural Networks (GNNs) with Recurrent Neural Networks (RNNs) to effectively learn a spatial-temporal representation of dynamic tactical networks. To the best of our knowledge, we are the first to propose a graph encoder-decoder framework for learning representations of tactical communication networks.
\item \textbf{Experimental Validation:} Through comprehensive experiments, we demonstrate the superior performance of STGED, achieving 99.2\% accuracy in predicting future network connectivity and outperforming baselines of recent works.
\end{itemize}

\section{Related Work}

\subsection{Tactical Communication Networks}

A considerable body of research has applied the Anglova scenario to study TCNs. Lopatka et al. \cite{lopatka_multi_2022} focused on spectrum optimization to mitigate congestion, while Suri et al. \cite{suri_forward_2022} improved Blue Force Data dissemination through optimized link state routing (OLSR) and simplified multicast forwarding. Addressing the importance of communication reliability, Schutz et al. \cite{schutz_network_2021} proposed a multi-path forward erasure correction method based on network coding. In contrast, Rechenberg et al. \cite{10253464} utilized recurrent neural networks (RNNs) for link prediction but did not consider tactical mobility models or terrain effects, which are critical in TCNs.

\subsection{Link Prediction in MANETs}

Link prediction has been widely studied across various domains, including MANETs, which share similarities with TCNs. Preetha et al. \cite{preetha_fuzzy_2018} used fuzzy decision processes and belief propagation to predict link duration. Patel et al. \cite{10.1504/ijhpcn.2020.112694} employed statistical techniques based on signal strength for the same purpose. Furthermore, Shao et al. \cite{shao_link_2022} combined graph convolutional networks (GCN) with gated recurrent units (GRU) to capture spatial-temporal features in dynamic networks.

\subsection{GNNs in Networking}

GNNs have shown promise in various networking applications. Maret et al. \cite{maret_investigation_2022} and Swaminathan et al. \cite{swaminathan_graphnet_2021} applied GNNs to optimize routing in the Anglova scenario and demonstrated generalizability across different network topologies. Sun et al. \cite{sun_combining_2021} optimized virtual network function placement using a GNN with reinforcement learning (RL). Applications extend to Internet of Things (IoT) networks, where GNNs have been used for optimal resource allocation \cite{chen_gnn-based_2022} and traffic prediction \cite{guo_traffic_2023}.

\subsection{Spatial-Temporal Graph Neural Networks (STGNN)}

Special-temporal models are widely used to model sequential applications in the natural language processing domain, such as itinerary planning~\cite{liu2020strategic}, social media mining~\cite{liu2021crisisbert,liu2022title2vec,liu2020ipod}, geo-tagging~\cite{9006284,
george2021real}, and other applications~\cite{singhal2021analyzing,heng2020urban,li2023transformer}. In networking, STGNNs are relevant for capturing data correlation in both time and space. For example, Chen et al. \cite{chen_novel_2023} proposed an STGNN combining GCN and long short-term memory (LSTM) unit for cellular network traffic prediction. Wang et al. \cite{wang_spatial-temporal_2023} outperformed traditional GNN and GRU models in predicting cellular traffic. In the context of road traffic forecasting, Lan et al. \cite{pmlr-v162-lan22a} introduced a dynamic STGNN that significantly outperformed state-of-the-art methods. Li et al. \cite{Li_Zhu_2021} proposed a fusion operation on multiple spatial and temporal graphs for synchronized learning of hidden spatial-temporal correlations. Li et al. \cite{li2023staged} proposed a graph encoder-decoder approach to conduct fault diagnosis in industrial processes.

\subsection{Summary}

The aforementioned works demonstrate significant advancements in TCNs, link prediction in MANETs, and the application of GNNs in various networking contexts. However, there is a gap in the literature concerning the effective incorporation of spatial-temporal modeling in TCNs for enhanced performance and reliability. Motivated by the exceptional performance of STGNNs in other domains, this paper aims to bridge this gap by applying advanced spatial-temporal modeling techniques to TCNs.
\begin{figure*}[t]
    \begin{center}
        \includegraphics[width=1.\linewidth]{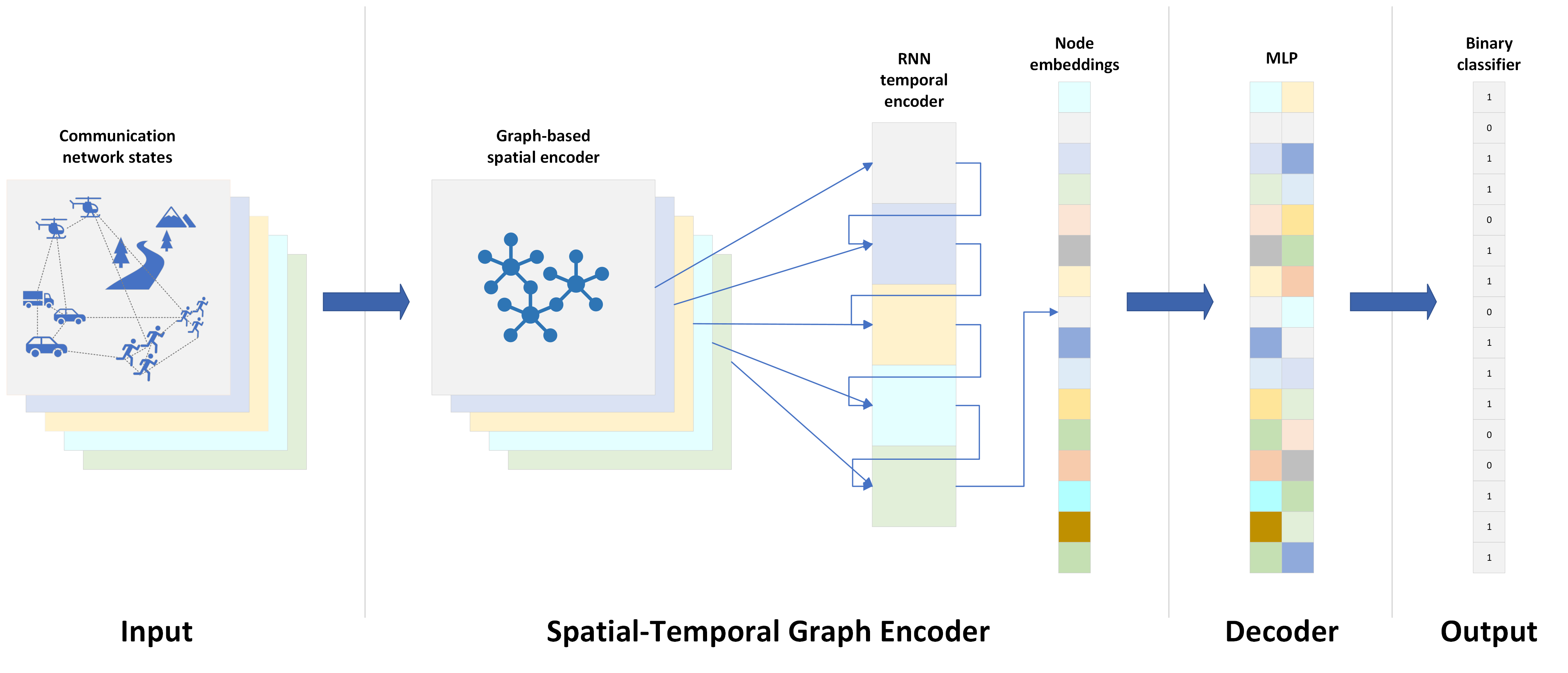}
    \end{center}
    \caption{A Spatial-Temporal Graph Encoder-Decoder (STGED) framework for modelling Tactical Communication Networks (TCN). A series of TCN states pass through a graph-based neural network to learn the spatial features for each state. Subsequently, the states form a time-series input and passes through a recurrent neural network (RNN) to learn the temporal features, i.e. the tactical movement patterns. The last hidden layer of the RNN acts as the embeddings of nodes in the TCN. The decoder takes pairs of node embeddings as input and predicts their connectivity at the next time step. }
    \label{fig:stged}
    \vspace{-0.5cm}
\end{figure*}
\section{Methodology}


\subsection{Preliminary}
We denote the Tactical Communication Network (TCN) as a temporal graph $G^{(t)}=(V^{(t)}, E^{(t)}, X_V^{(t)}, X_E^{(t)})$, where $V^{(t)} \in \mathbb{R}^N$ represents a finite set of $N$ nodes (i.e. communication devices) within the TCN; $E^{(t)} \in \mathbb{R}^{M}$ is a set of $M$ edges that represent valid communications between pairs of nodes; $X_V^{(t)} \in \mathbb{R}^{C_X}$ denotes the set of node features (i.e. velocity) at time $t$; and $ X_E^{(t)} \in \mathbb{R}^{C_W}$ is the TCN's connectivity features (i.e. distance, path loss, prop-delay and timestamp) at time $t$.

This work aims to predict the connectivity status $E_{t+1}$ at time $t+1$, i.e., whether a link exists for a given pair of node, based on a window of past observations of the network states, i.e., $G_{(t-w)...t}$. Our goal is to train a STGED model to predict the network connectivity, i.e. links among given pairs of nodes, in the next state, as follows:

\begin{equation}
\begin{aligned}
    E_{t+1} & = STGED[G^{(t-w):t}] \\
            & = STGED[V^{(t-w):t}, E^{(t-w):t}, X^{(t-w):t}, W^{(t-w):t}]
\end{aligned}
\label{algo:ModNet}
\end{equation}

\subsection{Spatial-Temporal Graph Encoder-Decoder Framework}

We present the framework of Spatial-Temporal Graph Encoder-Decoder (STGED) for TCNs in Figure~\ref{fig:stged}. The Encoder-Decoder architecture is widely used in the Computer Vision (CV) and Natural Language Processing (NLP) communities for sequence classification tasks, such as machine translation and image captioning. However, as the features of TCN data are physical measures and do not contain explicit semantics like that in CV and NLP, learning a feature representation for the latent tactical behavior becomes more challenging. Therefore, we propose the Spatial-Temporal Graph Encoder-Decoder (STGED) framework to address this challenge. Intuitively, we first use a graph-based convolution network to learn the spatial representation of each network state. Subsequently, these state representations are passed as time-series inputs to a recurrent neural network to capture latent temporal features emerging from the evolution of the input states. 

Formally, the Encoder comprises two main components: a graph-based encoder and an recurrent temporal encoder. These components work together to learn a network embedding $Z_{(t-w):t}$ based on the node and edge features, as follows:

\begin{equation}
\begin{aligned}
    Z^{(t-w):t} = TE[SE[G^{(t-w):t}]]
\end{aligned}
\label{algo:encoder}
\end{equation}

The Decoder then takes in the spatial-temporal network embedding and predict the network connectivity at $t+1$:

\begin{equation}
\begin{aligned}
    E^{(t+1)} = DE[Z^{(t-w):t}]
\end{aligned}
\label{algo:decoder}
\end{equation}

\subsection{Spatial-Temporal Graph Encoder}
In TCN, both the communicating devices and their connectivity display spatial relationship in the network, such as the relative velocities and distance among connected nodes. We adopt a Graph Transformer Convolutional (GTC) unit~\cite{shi2021masked} that uses multi-head attention mechanism to compute the latent relationship among all the nodes. Concretely, the GTC unit computes the node representation $x^{\prime}_i$ of $i$th node where $i < N$ by propagating information from the $i$th node and all other $j$th nodes $j < N$, $j\neq i$, thereby learning the global latent temporal features:

\begin{equation}
\begin{aligned}
    x^{\prime}_i = \mathbf{W}_1 \mathbf{x}_i+\sum_{j \in \mathcal{N}(i)} \alpha_{i, j}\left(\mathbf{W}_2 \mathbf{x}_j+\mathbf{W}_3 \mathbf{e}_{i j}\right)
\end{aligned}
\label{algo:SE}
\end{equation}

and attention coefficients $\alpha_{i, j}$ as:
\begin{equation}
\begin{aligned}
\alpha_{i, j}=\operatorname{softmax}(\frac{(\mathbf{W}_4 \mathbf{x}_i)(\mathbf{W}_5 \mathbf{x}_j+\mathbf{W}_6 \mathbf{e}_{i j})}{\sqrt{d}})
\end{aligned}
\label{algo:atcoef}
\end{equation}
where $x_i, x_j \in X_E$, edge features $e_{ij} \in X_V$, $\sqrt{d}$ as a scaling factor for $d$ dimensions, and $\mathbf{W_{1:6}}$ are trainable parameters.

After that, a series of the GTC units from time $t-w$ to $t$ will then be passed into a LSTM to learn the \textit{temporal} behaviours from the evolution of network states:

\begin{equation}
\mathbf{h}_t, \mathbf{c}_t = \text{LSTM}(\mathbf{W}, \mathbf{b}, \mathbf{h}_{t-1}, \mathbf{c}_{t-1}, \mathbf{x}_t)
\label{algo:lstm}
\end{equation}

where \( \mathbf{W} = \{\mathbf{W}_f, \mathbf{W}_i, \mathbf{W}_o, \mathbf{W}_g\} \) are trainable weight metrics, \( \mathbf{b} = \{\mathbf{b}_f, \mathbf{b}_i, \mathbf{b}_o, \mathbf{b}_g\} \) are trainable bias factors, for the LSTM's forget, input, output, and candidate cell state, respectively. The LSTM calculates the new hidden state $h_t$ and cell state $c_t$ previous hidden state \( \mathbf{h}_{t-1} \) by previous cell state \( \mathbf{c}_{t-1} \), \( \mathbf{x}^{\prime}_t \).

Finally, we take the last hidden state of the LSTM as the TCN's state encoder from time $t-w$ to $t$, i.e.:
\begin{equation}
Z^{(t-w):t} = \mathbf{h}_t
\label{algo:hiddenstate}
\end{equation}

We set $Z^{(t-w):t} \in \mathbb{R}^{N \times H}$, where $N$ is the number of nodes and $H$ is the node embedding size, which is a tune-able hyperparameter. This implicitly instructs the STGE to learn effective embeddings of each node. As a result, we could simply pass embeddings of two nodes into the decoder to predict their connectivity at a future state.

\subsection{Multi-Layer Perceptron Decoder}

The decoder module is designed to predict the connectivity between pairs of nodes in the graph. Specifically, it takes as input the node embeddings generated by the LSTM encoder and outputs a connectivity score for each pair of nodes. Concretely, the decoder can be formulated as follows:

Let \( \mathbf{emb}^{(i)} \) and \( \mathbf{emb}^{(j)} \) be the embeddings for nodes \( i \) and \( j \) respectively, where \( \mathbf{emb}^{(i)}, \mathbf{emb}^{(j)} \in \mathbb{R}^H \) and \( H \) is the dimension of node embeddings. 

These vectors are then passed through the MLP decoder to obtain the connectivity score \( s_{ij} \):

\begin{equation}
s_{ij} = \text{MLP}(\mathbf{W}, \mathbf{b}, \mathbf{Em})
\end{equation}

where $\mathbf{Em}=concat(\mathbf{emb}^{(i)}, \mathbf{emb}^{(j)})$.

Here, \( \text{MLP} \) is a function representing the Multi-Layer Perceptron, parameterized by weight matrices \( \mathbf{W} \) and bias vectors \( \mathbf{b} \). The function takes the concatenated node embeddings \( \mathbf{z}_{ij} \) as input and outputs a scalar \( s_{ij} \) that represents the predicted connectivity between nodes \( i \) and \( j \).

At training time, the output \( s_{ij} \) is kept as the probability of the connectivity that will then be passed into a loss function. At testing time, the score is then thresholded to make a binary decision about whether nodes \( i \) and \( j \) are connected or not. Specifically, if \( s_{ij} \) is greater than a predefined threshold \( \tau \) the nodes are considered to be connected. 

\begin{equation}
\text{Connectivity}_{ij} = 
\begin{cases} 
1 & \text{if } s_{ij} > \tau \\
0 & \text{otherwise}
\end{cases}
\end{equation}

To allow the decoder module to effectively map the high-dimensional embeddings to a connectivity prediction, we perform next-state one-hop connectivity prediction task with a Binary Cross Entropy (BCE) loss function with $\tau=0.5$\footnote{Empirically, the convergence rate and performance of the model are not sensitive to the choice of the threshold.}. 
\begin{figure*}[t]
     \centering
     \begin{subfigure}[b]{0.195\textwidth}
         \centering
         \includegraphics[width=0.95\textwidth]{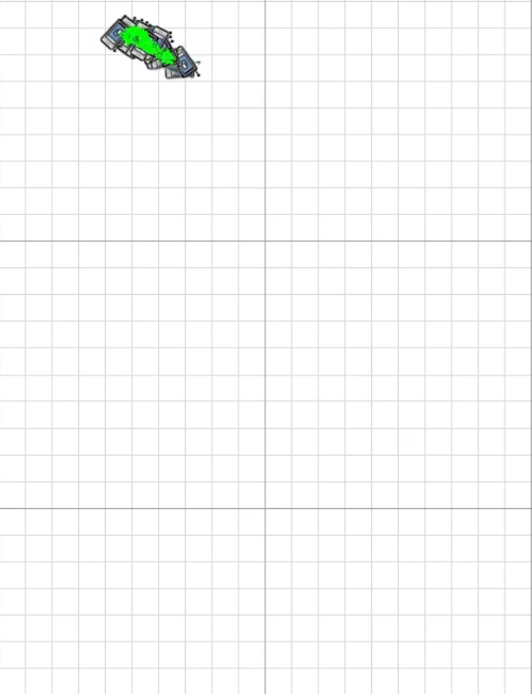}
         \caption{CNTM t=0m}
         \label{subfig:anglova0}
         \vspace*{2mm}
     \end{subfigure}
     \begin{subfigure}[b]{0.195\textwidth}
         \centering
         \includegraphics[width=0.95\textwidth]{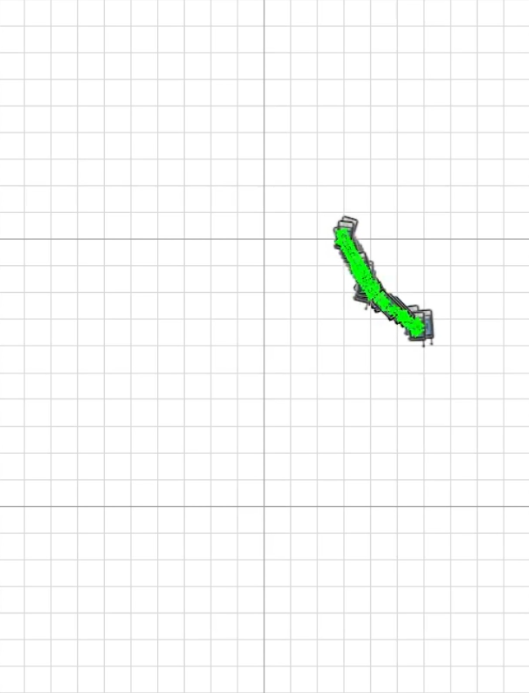}
         \caption{CNTM t=30m}
         \label{subfig:anglova30}
         \vspace*{2mm}
     \end{subfigure}
     \begin{subfigure}[b]{0.195\textwidth}
         \centering
         \includegraphics[width=0.95\textwidth]{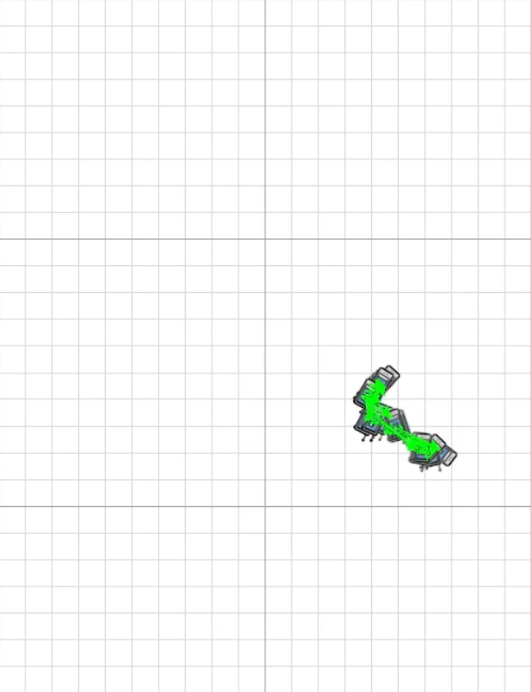}
         \caption{CNTM t=60m}
         \label{subfig:anglova60}
         \vspace*{2mm}
     \end{subfigure}
    \begin{subfigure}[b]{0.195\textwidth}
         \centering
         \includegraphics[width=0.95\textwidth]{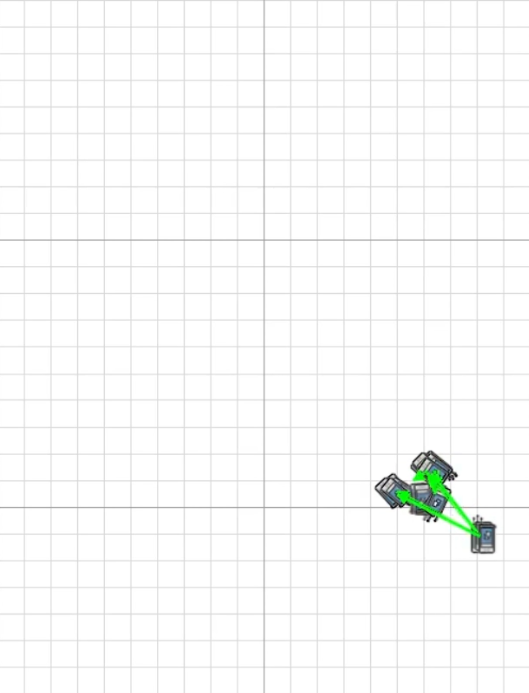}
         \caption{CNTM t=90m}
         \label{subfig:anglova90}
         \vspace*{2mm}
     \end{subfigure}
    \begin{subfigure}[b]{0.195\textwidth}
         \centering
         \includegraphics[width=0.95\textwidth]{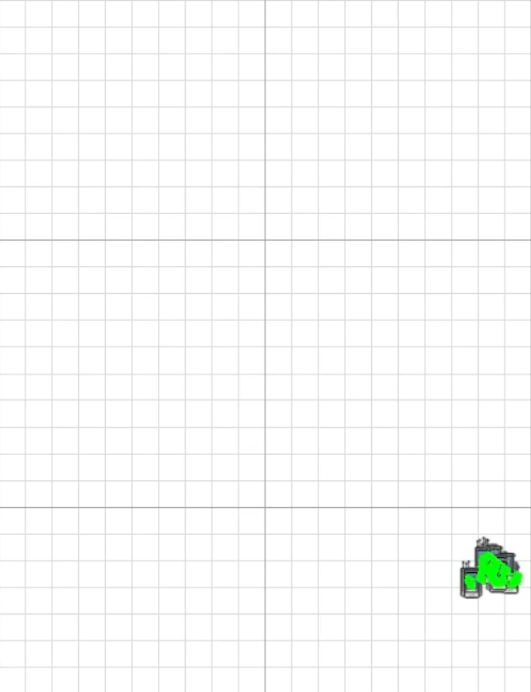}
         \caption{CNTM t=120m}
         \label{subfig:anglova120}
         \vspace*{2mm}
     \end{subfigure}
     \begin{subfigure}[b]{0.195\textwidth}
         \centering
         \includegraphics[width=0.95\textwidth]{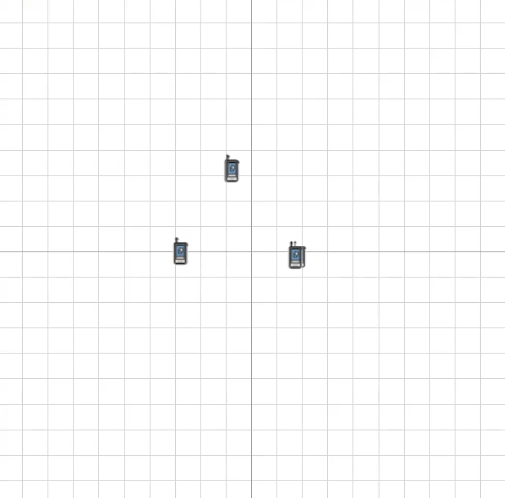}
         \caption{CNCM t=0m}
         \label{subfig:casual0}
     \end{subfigure}
     \begin{subfigure}[b]{0.195\textwidth}
         \centering
         \includegraphics[width=0.95\textwidth]{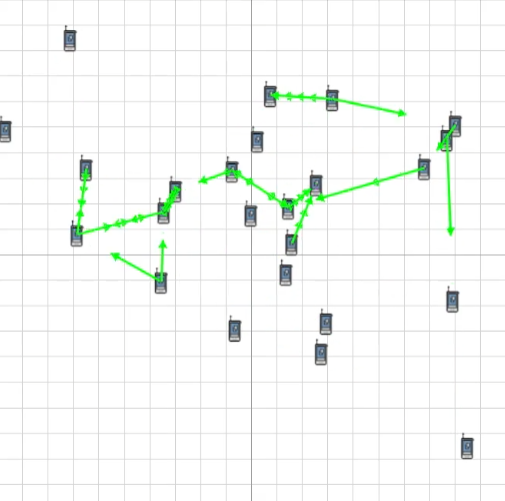}
         \caption{CNCM t=30m}
         \label{subfig:casual30}
     \end{subfigure}
     \begin{subfigure}[b]{0.195\textwidth}
         \centering
         \includegraphics[width=0.95\textwidth]{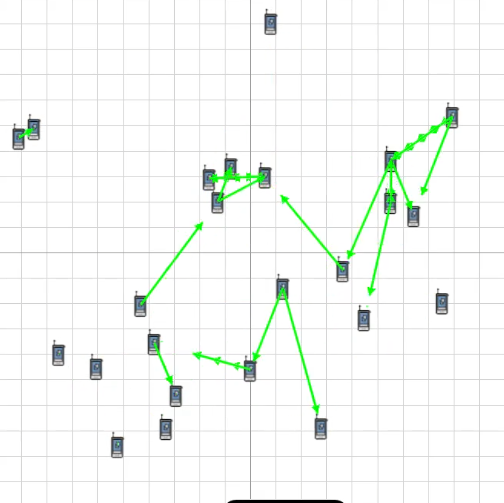}
         \caption{CNCM t=60m}
         \label{subfig:casual60}
     \end{subfigure}
    \begin{subfigure}[b]{0.195\textwidth}
         \centering
         \includegraphics[width=0.95\textwidth]{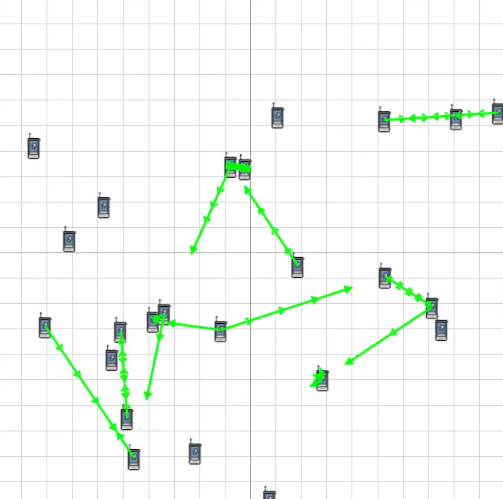}
         \caption{CNCM t=90m}
         \label{subfig:casual90}
     \end{subfigure}
    \begin{subfigure}[b]{0.195\textwidth}
         \centering
         \includegraphics[width=0.95\textwidth]{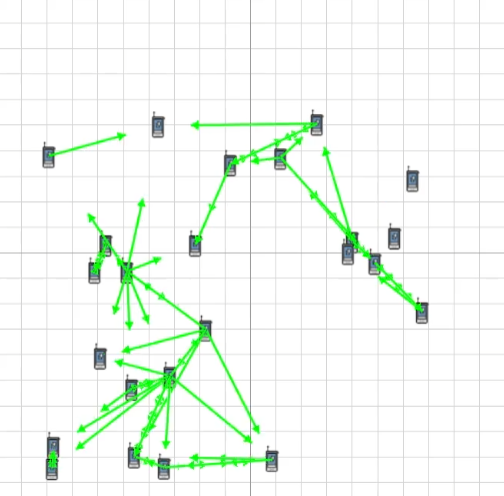}
         \caption{CNCM t=120m}
         \label{subfig:casual120}
     \end{subfigure}
\caption{Node positions over time (in minutes) for CNTM and CNCM datasets. The top row shows the objective-oriented mobility of the CNTM dataset, the tank company deploys from the upper left to bottom right. The bottom row shows the random mobility of the CNCM dataset. Nodes start off in 3 groups and randomly move toward and away from each other.}
\label{fig:CNTM_CNCM_mobility}
\vspace{-0.3cm}
\end{figure*}

\begin{figure}[t]
     \centering
     \includegraphics[height=4cm,width=8cm]{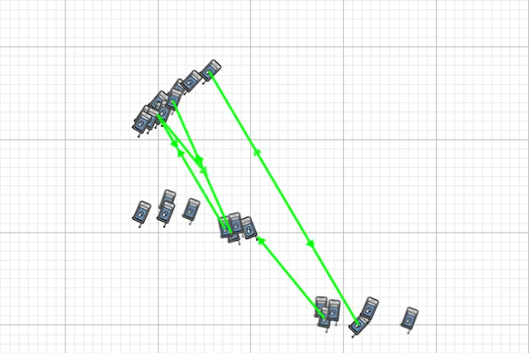}
     \caption{CNTM t=60m magnified}
     \label{fig:anglova60zoom}
     \vspace{-0.6cm}
\end{figure}

\begin{figure*}[t]
     \centering
     \begin{subfigure}[b]{0.195\textwidth}
         \centering
         \includegraphics[width=0.95\textwidth]{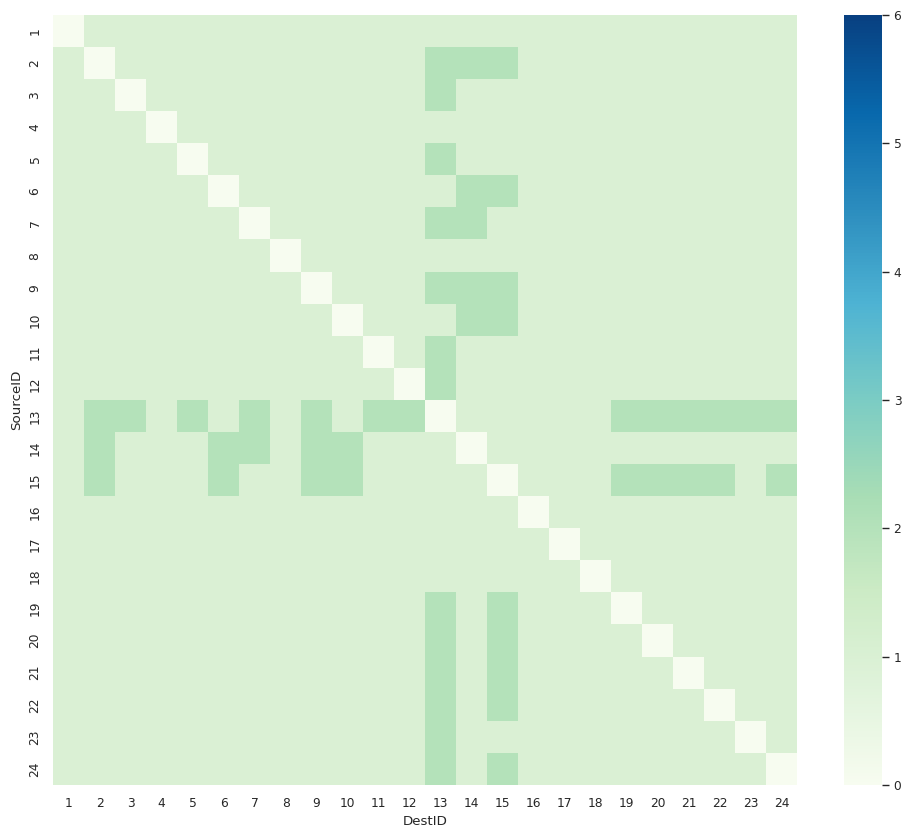}
         \caption{CNTM t=0m}
         \label{subfig:anglova0hop}
         \vspace*{2mm}
     \end{subfigure}
     \begin{subfigure}[b]{0.195\textwidth}
         \centering
         \includegraphics[width=0.95\textwidth]{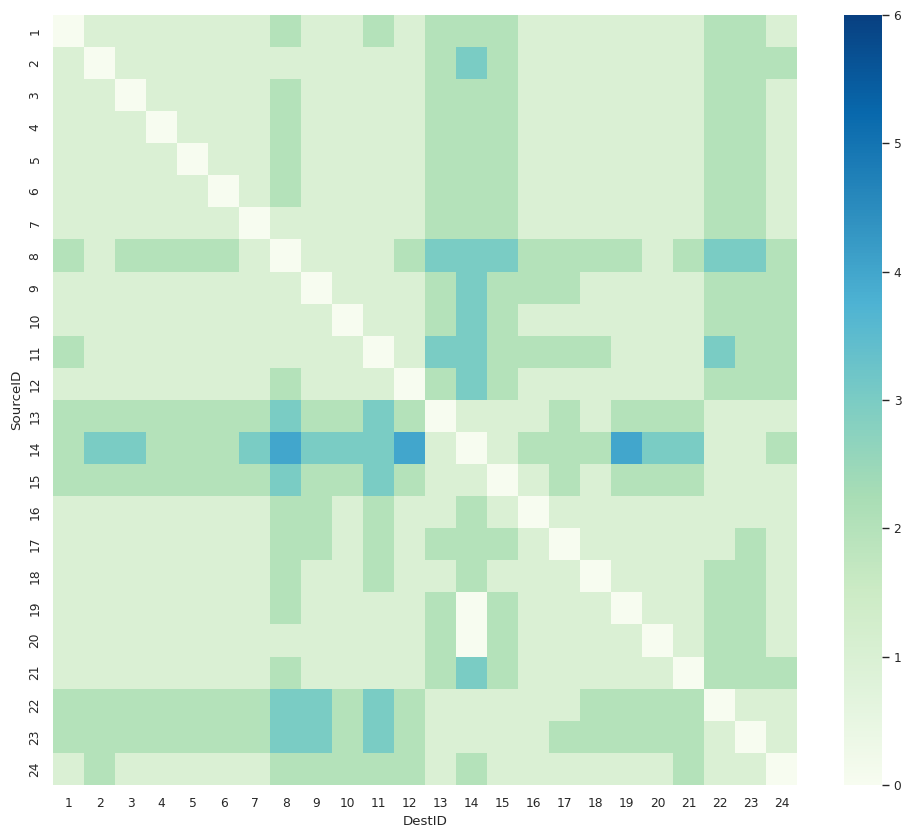}
         \caption{CNTM t=30m}
         \label{subfig:anglova30hop}
         \vspace*{2mm}
     \end{subfigure}
     \begin{subfigure}[b]{0.195\textwidth}
         \centering
         \includegraphics[width=0.95\textwidth]{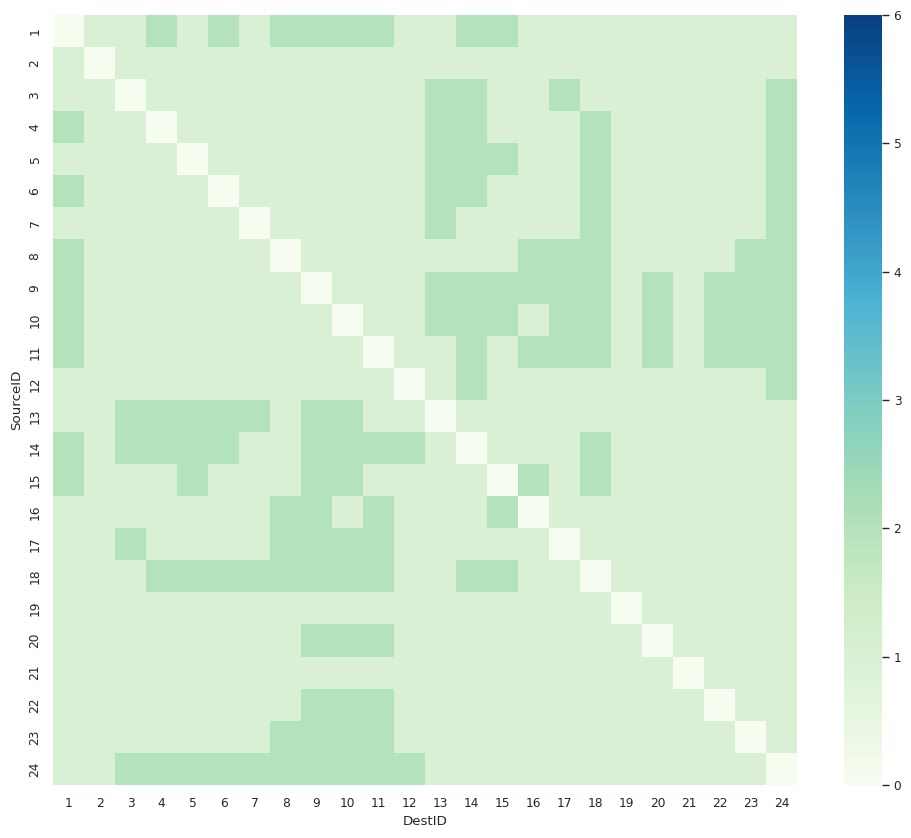}
         \caption{CNTM t=60m}
         \label{subfig:anglova60hop}
         \vspace*{2mm}
     \end{subfigure}
    \begin{subfigure}[b]{0.195\textwidth}
         \centering
         \includegraphics[width=0.95\textwidth]{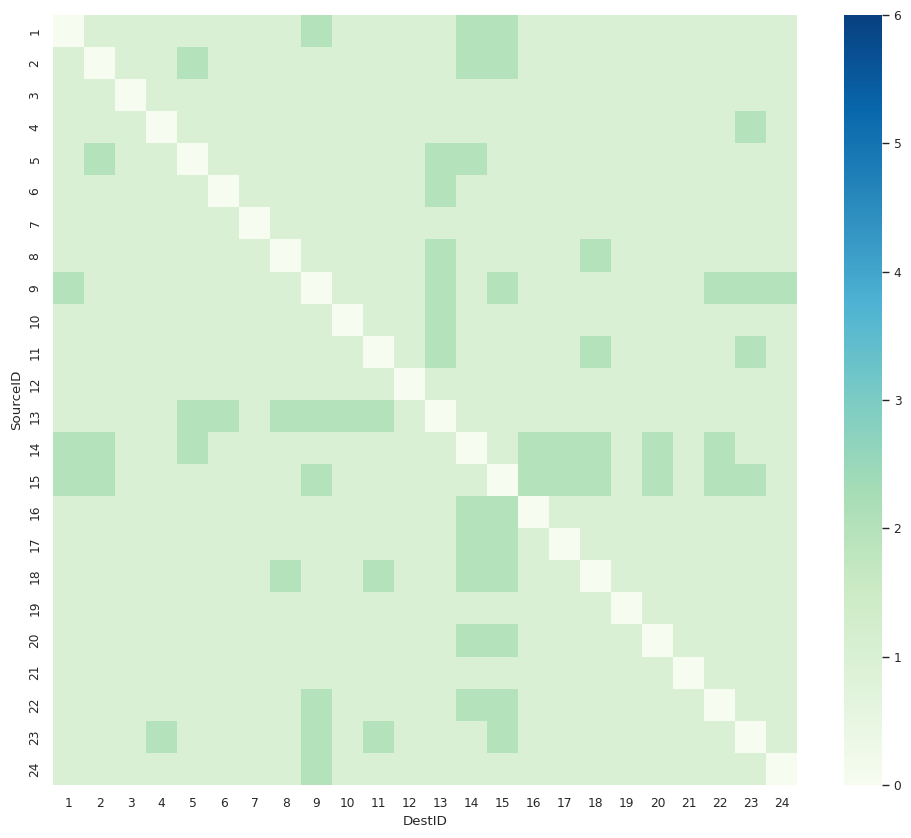}
         \caption{CNTM t=90m}
         \label{subfig:anglova90hop}
         \vspace*{2mm}
     \end{subfigure}
    \begin{subfigure}[b]{0.195\textwidth}
         \centering
         \includegraphics[width=0.95\textwidth]{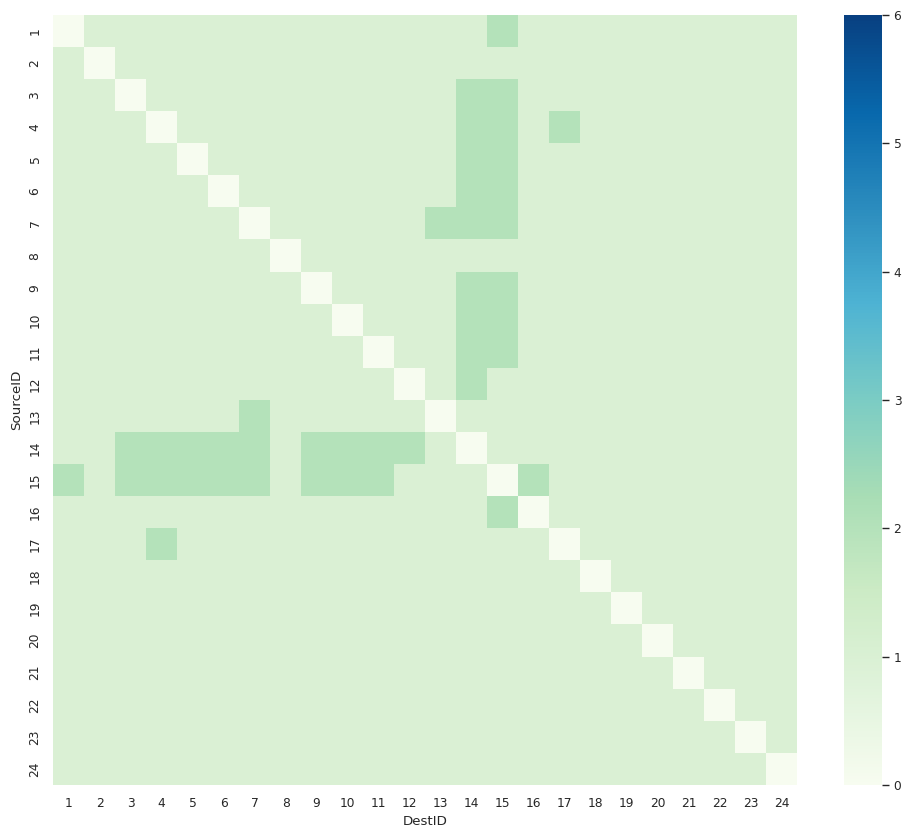}
         \caption{CNTM t=120m}
         \label{subfig:anglova120hop}
         \vspace*{2mm}
     \end{subfigure}
     \begin{subfigure}[b]{0.195\textwidth}
         \centering
         \includegraphics[width=0.95\textwidth]{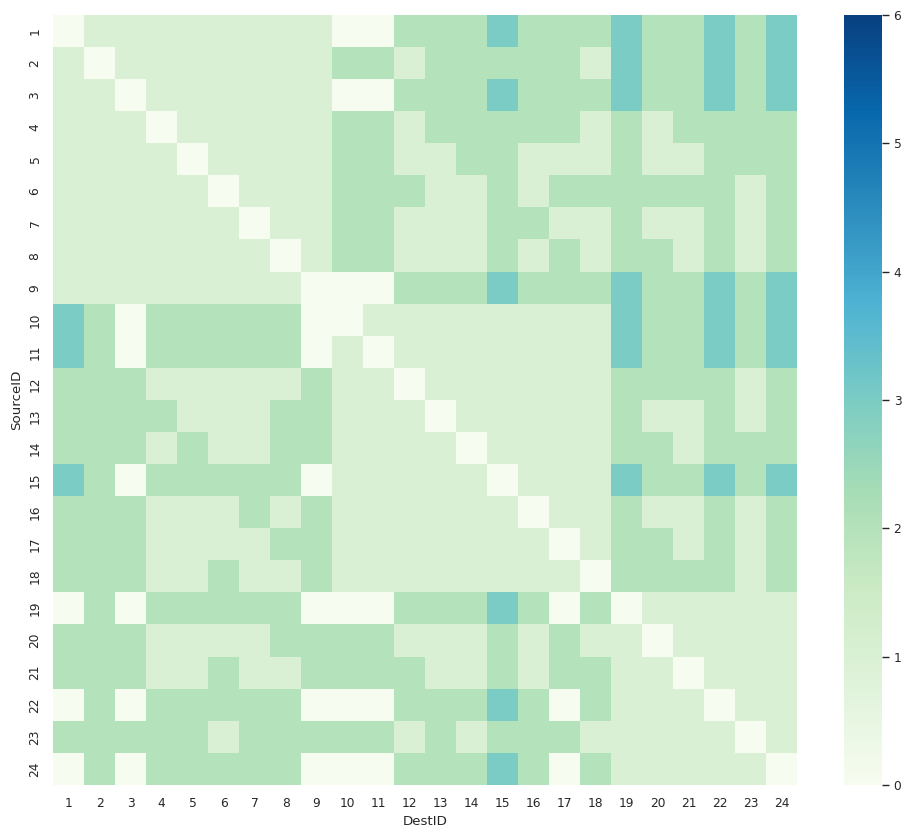}
         \caption{CNCM t=0m}
         \label{subfig:casual0hop}
     \end{subfigure}
     \begin{subfigure}[b]{0.195\textwidth}
         \centering
         \includegraphics[width=0.95\textwidth]{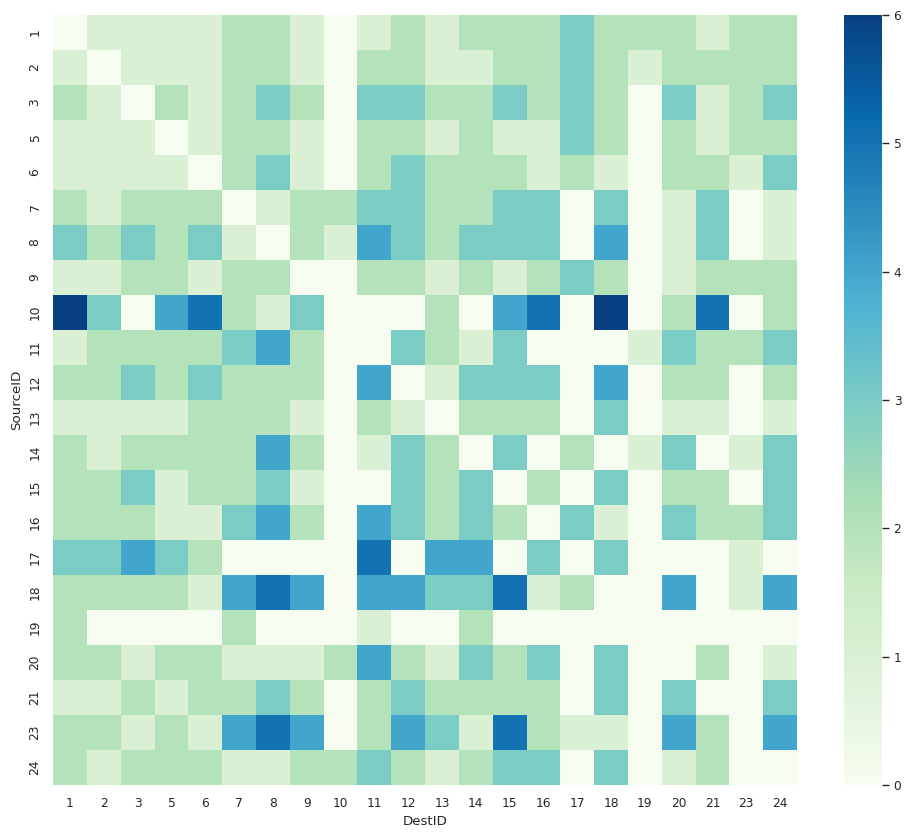}
         \caption{CNCM t=30m}
         \label{subfig:casual30hop}
     \end{subfigure}
     \begin{subfigure}[b]{0.195\textwidth}
         \centering
         \includegraphics[width=0.95\textwidth]{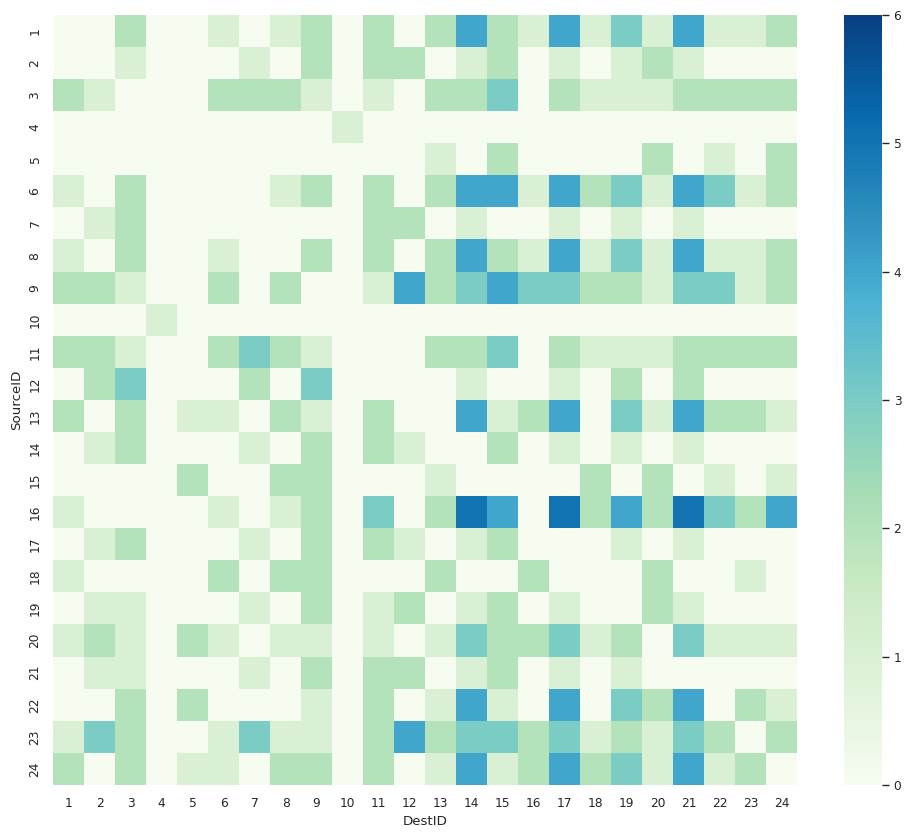}
         \caption{CNCM t=60m}
         \label{subfig:casual60hop}
     \end{subfigure}
    \begin{subfigure}[b]{0.195\textwidth}
         \centering
         \includegraphics[width=0.95\textwidth]{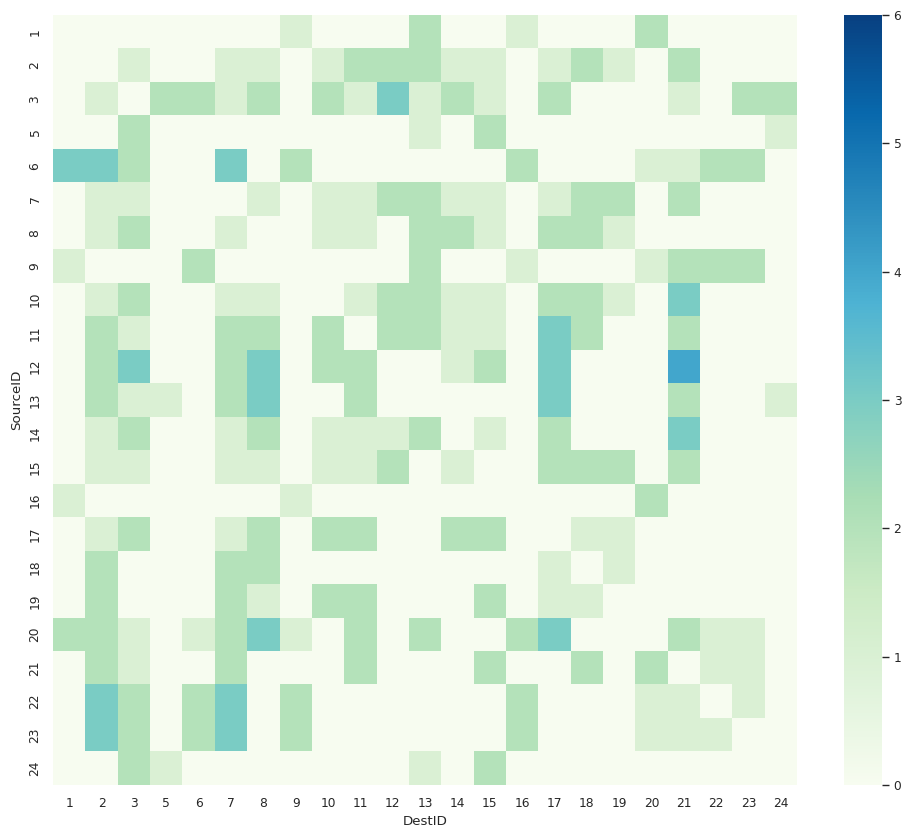}
         \caption{CNCM t=90m}
         \label{subfig:casual90hop}
     \end{subfigure}
    \begin{subfigure}[b]{0.195\textwidth}
         \centering
         \includegraphics[width=0.95\textwidth]{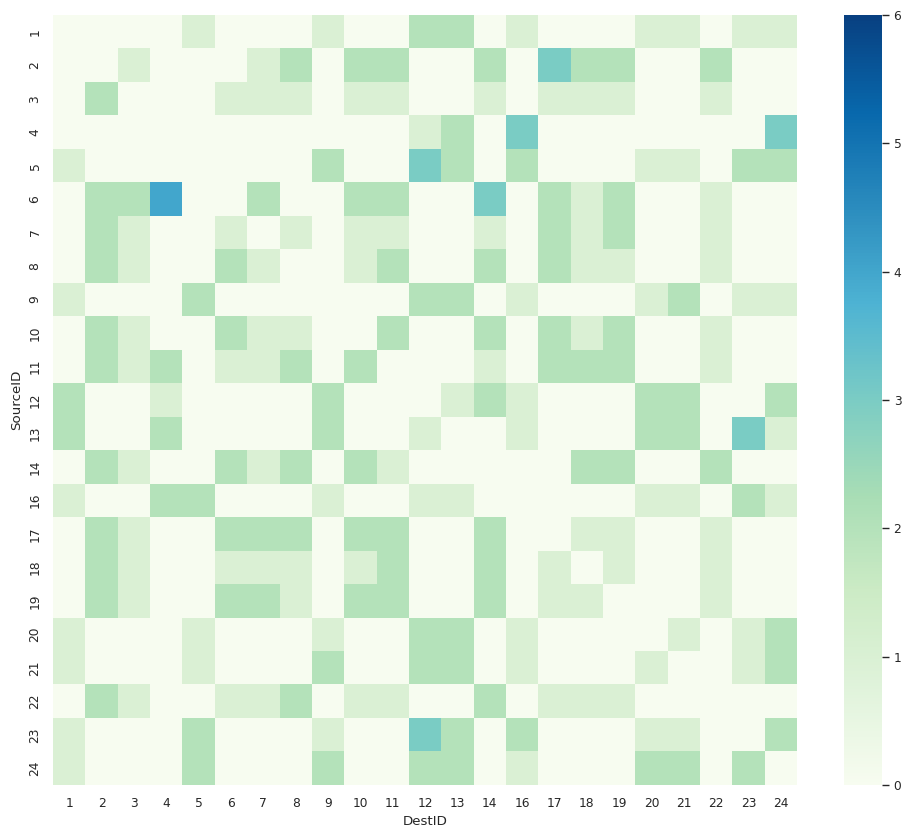}
         \caption{CNCM t=120m}
         \label{subfig:casual120hop}
     \end{subfigure}
\caption{Node hop count over time for the CNTM and CNCM datasets. This refers to the amount of nodes needed to transmit from the source node (y-axis) to the destination node (x-axis). If a node can send to another node directly, the hop count is 1. If a node has no route to another node, the hop count is 0 and the color is white. The top row of figures shows the number of hops needed to connect nodes of the CNTM dataset, and has a maximum of 4 hops at t=30m. The bottom row of figures illustrates number of hops needed to connect nodes of the CNCM dataset, and has a maximum of 6 hops at t=30m.}
\label{fig:CNTM_CNCM_hops}
\end{figure*}

\section{Experiments}

\subsection{Datasets}
For our TCN experiments we generated datasets using EXata \cite{keysight_exata_nodate}, a high fidelity network emulation toolset. EXata uses a software virtual network to represent networks across various protocol levels and radio types, and can even interoperate with real devices to provide hardware-in-the-loop capabilities. We used EXata to model a subset of the Anglova scenario. Table~\ref{tab:dataset_stat} summarizes the statistics of both datasets.

\begin{table}[h]
    \centering
    \scalebox{1.3}{
    \begin{tabular}{|c|c|c|}
        \hline
         & CNTM& CNCM\\
        \hline
        \#States & 3998& 3599\\
        \hline
        Avg. Nodes & 24.0 & 24.0 \\
        \hline
        Avg. Edges & 685.20& 1280.55 \\
        \hline
        Node features & \multicolumn{2}{c|}{ VelocityX, VelocityY}  \\
        \hline
        \multirow{2}{*}{Edge features}  & \multicolumn{2}{c|}{Distance, Path loss,}  \\
                                    & \multicolumn{2}{c|}{PropDelay, Timestamp} \\                      
        \hline
    \end{tabular}}
    \caption{Statistics of CNTM and CNCM datasets. Note that the Avg. Edges are more than a fully-connected graph with 24 nodes in one-second interval. Within each time step, there could exist none, one or many edges between any two nodes.}
    \label{tab:dataset_stat}
    \vspace{-0.4cm}
\end{table}

\subsubsection{Communication Network with Tactical Movements (CNTM)} 
For our first dataset, named Communication Network with Tactical Movements (CNTM), we used the subset of the Anglova scenario including a single tank company using UHF radios during vignette 2, which is the troop deployment stage. For this stage we used node position information and a radio path loss matrix provided on the Anglova website \cite{peuhkuri_start_2022}. This data has a granularity of one second and a duration of 7200 seconds, or 2 hours. In EXata we configured the nodes to follow the provided node positions and use the provided path loss. We also configured the UHF radios, the time division multiple access (TDMA) protocol, and OLSRv2 routing protocol per the Anglova definitions \cite{8398729}. EXata can be configured to log events that occur at the physical layer, MAC layer, network layer, and application layer. In addition to the events, EXata also can regularly save status tables at user defined granularity. These events and status tables across the 2 hour simulation make up the dataset and are saved in SQLite databases for further processing. During the experiment, the majority of the traffic is generated by the OLSRv2 routing messages, HELLO messages that nodes broadcast every 2 seconds, and Topology Control (TC) messages every 8 seconds. This is enough to keep connectivity states updated at a minimum of every 2 seconds. In the future, additional tactical applications such as Blue Force Tracking will be added.

\subsubsection{Communication Network with Casual Movements (CNCM)}
The second dataset, named Communication Network with Casual Movements (CNCM), is setup identically to the CNTM dataset except for node positions, mobility, and path loss. The mobility uses a random waypoint model provided in EXata, with a maximum speed of 10 meters per second. Because this does not use the Anglova positions, we cannot use the Anglova path loss values. Instead we use the two-ray ground reflection radio propagation model provided by EXata. This model is similar to a free space propagation model, with the addition of interference reflected from the ground. 

\subsubsection{Visualisation} 
As shown in Figure \ref{fig:CNTM_CNCM_mobility}, the random nature of the CNCM dataset and the objective-oriented mobility of the CNTM dataset. Note that the arrows between nodes represent messages being sent successfully. Nodes that are not connected cannot successfully send messages, but not all nodes that are connected will be sending messages at any given snapshot. Also note that the scale is different, as the CNTM travels a greater distance, but generally is grouped together. If the same scale was used for the CNCM dataset, there would be much fewer connections, which would skew the data. To help visualize the CNTM dataset, Figure \ref{fig:anglova60zoom} provides a magnified snapshot of t=60m. In this figure the nodes are on a path that is winding about a large hill, which blocks communication between many nodes on opposite sides. This is one example how the terrain may impact the connectivity more than the distance.

The node connectivity and number of hops between nodes is displayed in Figure \ref{fig:CNTM_CNCM_hops} for both the CNCM and CNTM datasets. We consider the number of hops as the minimum number of nodes required to send a message from one node to another, where nodes with direct connectivity are 1 hop away. In this figure, the x-axis refers to the destination node, and the y-axis refers to the source node. Note the diagonal line of white blocks in each subfigure where each node has no connectivity to itself. Additionally, there is some symmetry across this diagonal, since the path loss values are bidirectional. However, they are not perfectly symmetrical because the connectivity information updates when nodes send messages, and nodes do not send messages at the exact same time. 

In Figure \ref{subfig:anglova0hop}, we see most nodes in the CNTM dataset start off as 1-hop neighbors, while a few nodes are 2 hops away. As time progresses, the connectivity becomes more sparse in Figure \ref{subfig:anglova30hop}, with some nodes being 4 hops apart, and others without any connectivity. Similarly with the CNCM dataset,  the overall connectivity is highest in Figure \ref{subfig:casual0hop}, before becoming more sparse as nodes move away from each other in later time steps. The differences between the two datasets are represented in Figure \ref{fig:CNTM_CNCM_hops}, where the CNTM connections generally are more stable among groups of nodes. Nodes in CNTM are grouped in platoons and move in a coordinated manner, but these relationships are nonexistent for the CNCM dataset.

\subsubsection{Data Processing} 
For both datasets, we split the data into varying time steps, namely, one-, two- and five-steps, as input. We label the connectivity by existence of the sender-receiver pairs, and with an observed threshold of 128 dBm in path loss, i.e., a pair of nodes are connected if their path loss is 128 dBm or below. For training input of non-graph models, we concatenate the sender-node features, receiver-node features and the edge features into one vector without pooling or compression. 

\begin{table}[h]
\centering
\begin{tabular}{|c|c|c|}
\hline
Model & Architecture & Encoding \\
\hline
MLP & non-graph & general \\
LSTM & non-graph & temporal \\
GRU & non-graph & temporal \\
\hline
GCN & graph-based & spatial \\
GAT & graph-based & temporal \\
GATv2 & graph-based & temporal \\
GTC & graph-based & temporal \\
\hline
\end{tabular}
\caption{Model Categorization}
\label{tab:model_categorization}
\vspace{-0.5cm}
\end{table}

\subsection{Baselines}
We compare our proposed STGED framework with the following baselines:
\begin{itemize}
    \item \textbf{MLP}~\cite{haykin1994neural}: Multi-Layer Perceptron, which is a fully connected multi-layer neural network.
    \item \textbf{LSTM}~\cite{hochreiter1997long}: Long Short-Term Memory, which is a recurrent neural network.
    \item \textbf{GRU}~\cite{chung2014empirical}: Gated Recurrent Unit, which is a recurrent neural network.
    \item \textbf{GCN}~\cite{gilmer2017neural}: Graph Convolutional Network with Edge-Conditioned convolutional filters.
    \item \textbf{GAT}~\cite{velickovic2017graph}: Graph Attention Network where its attention mechanism uses a single-layer feed-forward neural network.
    \item \textbf{GATv2}~\cite{brody2021attentive}: Graph Attention Network with improvement, where every node can attend to any other node.
    \item \textbf{GTC}~\cite{shi2021masked}: Graph Transformer Convolutional network that uses scaled dot-product attention, similar to the original Transformer model.
\end{itemize}

Table~\ref{tab:model_categorization} summarises the categories of the models in architectural and encoding types.

\subsection{Experiment Setup}
We split the datasets into 90:5:5 ratio for training, validation and test sets, respectively. As the baselines contains non-temporal models, we choose to split the data randomly instead of chronologically for fair evaluation. 

To balance the positive and negative samples, we take the minimum of positive and negative sample sets, and randomly sample the same number of samples from the larger set in training time. In testing time, we retain all positive and negative samples for performance evaluation. We apply a 20\% dropout to all components to further prevent over-fitting.


All models are trained on the CNTM dataset. The datasets are processed to have three different sizes of time steps, i.e., one-, two-, and five-step, as the input to the models to predict the next-state connectivity of the GTC. We measure the performance of the models with four metrics, namely accuracy (Acc), Precision (Pre), Recall (Rec) and F1-score (F1). 

In terms of hyperparameters for STGED, the spatial encoder has two hidden layers with 1024 hidden units and 128 attention heads. The node embedding is of same size as the hidden units, i.e. 1024. The temporal encoder (i.e. LSTM) has two hidden layers and 2048 hidden units, which do not process compression to the two input node embeddings. All models runs over 20 epochs with a learning rate of 1e-6, and with additional epochs until they converge. 

For fair comparison, all baselines are constructed with similar scale and trained with similar hyperparameters as compared to the proposed models.

All experiments are conducted on an Nvidia DGX server with 4 Tesla V100-DGXS-32GB GPUs. Each training or testing is conducted on a single GPU at a time. 

\begin{table*}[t]
\centering
\scalebox{1.2}{
\begin{tabular}{|c|cccc|cccc|cccc|}
\hline         
\multirow{2}{*}{} & \multicolumn{4}{c|}{1 Step} & \multicolumn{4}{c|}{2 Steps} & \multicolumn{4}{c|}{5 Steps} \\
\cline{2-13}                         
                 & Acc & F1 & Pre & Rec  & Acc & F1 & Pre & Rec & Acc & F1 & Pre & Rec  \\
\hline
MLP              & 0.868  & 0.921  & 0.929  & 0.938 & 0.910 & 0.947  & 0.970  & 0.995 & 0.920 & 0.935  & 0.943  & 0.951  \\
GRU              & 0.783  & 0.855  & 0.835  & 0.815 & 0.893  & 0.915  & 0.955  & 0.998 & 0.918 & 0.925  & 0.959  & 0.996 \\
LSTM             & 0.783  & 0.870  & 0.869  & 0.869 & 0.929  & 0.956  & 0.976  & 0.996 & 0.933 & 0.957  & 0.976  & 0.996 \\
\hline
GCN              & 0.664  & 0.794  & 0.908 & 0.705  & 0.895  & 0.945  & 0.916  & 0.975 & 0.906  & 0.931  & 0.962  & 0.995 \\
GAT              & 0.869  & 0.945  & 0.930  & 0.916 & 0.901  & 0.934  & 0.965  & 0.998 & 0.918  & 0.957  & 0.918  & 1.0 \\
GATv2            & 0.906  & 0.951  & 0.956  & 0.953 & 0.916  & 0.956  & 0.978  & 1.0 & 0.936  & 0.985  & 0.992  & 1.0 \\
GTC              & 0.906  & 0.951  & 0.956  & 0.953 & 0.916  & 0.956  & 0.978  & 1.0 & 0.936  & 0.985  & 0.992  & 1.0 \\
\hline
\textbf{STGED}            & \textbf{0.919}  & \textbf{0.966}  & \textbf{0.982}  & \textbf{0.968} & \textbf{0.979}  & \textbf{0.991}  & \textbf{0.996}  & \textbf{1.0} & \textbf{0.992}  & \textbf{0.998}  & \textbf{0.998}  & \textbf{1.0} \\
\hline
\end{tabular}}
\caption{Experimental Results for STGED and baselines for the next-state 1-hop connectivity prediction. Using input time steps of one, two and five, STGED consistently outperforms all baseline models across all steps and metrics.}
\label{tab:exp_results}
\vspace{-0.3cm}
\end{table*}
\subsection{Results}

Table~\ref{tab:exp_results} shows the experimental results of the proposed STGED model and the six baseline methods. Below summarizes our observations and interpretations:
\begin{enumerate}
    \item \textbf{Strong Performance of STGED}: This model consistently outperforms all baseline models across all steps and metrics. It exhibits both high precision and high recall, suggesting excellent performance in both identifying and not missing positive instances. Furthermore, adding spatial encoding unit significantly improves the performance of the model across all steps.
    \item \textbf{High Consistency in Attention-Based Graph Models}: GAT, GATv2 and TCN exhibit very consistent performance metrics across different steps. This implies that the attention mechanism is effective in learning temporal feature representation for this task.
    \item \textbf{High Recall Across Models}: Almost all models, especially at two and five steps, have high or perfect recall. This is a significant trend that points to the models' ability to capture most of the positive instances. The high recall values could also indicate that the connectivity prediction task or the dataset makes it easier to capture positive instances. Given that different types of encoders (temporal, spatial, and general) also achieve high recall, this seems more a characteristic of the dataset or the task itself rather than the models.
\end{enumerate}

\begin{table}[t]
\centering
\label{tab:experimental_results}
\scalebox{1}{
\begin{tabular}{|c|cc|}
\hline
Encoder & Accuracy & F1-score \\
\hline
GCN              & 0.503  & 0.780 \\
GAT              & 0.508  & 0.423 \\
GATv2            & 0.501  & 0.484 \\
GTC              & 0.538 &  0.647 \\
\hline
GCN-GRU              & 0.686  & 0.780 \\
GAT-GRU              & 0.701  & 0.608 \\
GATv2-GRU            & 0.713  & 0.685 \\
GTC-GRU               & 0.720  & 0.682 \\
\hline
GCN-LSTM              & 0.686  & 0.580 \\
GAT-LSTM              & 0.708  & 0.623 \\
GATv2-LSTM            & 0.723  & 0.668 \\
\textbf{GTC-LSTM}     & \textbf{0.739}  & \textbf{0.720} \\
\hline
\end{tabular}}
\caption{Ablation Experimental Results}
\vspace{-0.6cm}
\label{tab:ablation}
\end{table}

\subsection{Ablation Experiments}
We conduct ablation experiment on proposed STGED framework to variate each component. For spatial encoder, we choose from: (1) GCN, (2) GAT, (3) GATv2 and (4) GTC. For temporal encoder, we choose between LSTM and GRU. Furthermore, we conduct further experimentations to assess effectiveness of temporal-encoder-only and spatial-encoder-only models. Finally, we use the CNCM dataset, i.e., random movements of the nodes without a goal, to benchmark the effectiveness in representation learning. All models use five time steps as input to predict the 1-hop connectivity of the following state.

Table \ref{tab:ablation} shows the results of the ablation experiment. We observe that GTC-LSTM shows the highest accuracy and F1-score. This aligns with our expectation because GTC uses dot-product attention among connected nodes, which has a global vision of latent spatial features; as for the temporal encoder, LSTM consists of more gates (i.e. input, output and forget gates) than the GRU (i.e. reset and update gates). While GRU is time-efficient in training, LSTM is more powerful and adaptable as a temporal encoder. 

\section{Conclusion}
This paper presents a novel STGED framework for TCN representation learning. Two datasets demonstrate the ability of the STGED in learning latent features, such as the spatial-temporal tactical movements. We design experiment tasks to pass through varying network states (i.e. time steps) to the models and predict the next-state one-hop connectivity. Experimental results show that the STGED consistently outperforms all baseline models across all input steps and metrics, revealing advantages in STGED for representing TCNs.
\section*{Acknowledgement}
This research is supported by the National Research Foundation, Singapore and Infocomm Media Development Authority under its Future Communications and Research \& Development Programme.

\bibliographystyle{IEEEtran}
\bibliography{IEEE}

\end{document}